\documentclass[runningheads]{llncs}
\usepackage{graphicx}
\usepackage{amsmath,amssymb} 
\usepackage{color}
\usepackage{subcaption}
\usepackage[width=122mm,left=12mm,paperwidth=146mm,height=193mm,top=12mm,paperheight=217mm]{geometry}
\begin{document}





\title{GAN-based Virtual Re-Staining:  A Promising Solution for Whole Slide Image Analysis } 

\author{Zhaoyang Xu$^{\star }$, Xingru Huang$^{\star }$, Carlos Fern\'andez Moro$^{\dagger}$, B\'ela Boz\'oky$^{\dagger}$, Qianni Zhang$^{\star }$ }
\institute{email: zhaoyang.xu@qmul.ac.uk.\\
$^{\star }$Queen Mary University of London, $^{\dagger}$Karolinska University Hospital}

\maketitle

\begin{abstract}
Histopathological cancer diagnosis is based on visual examination of stained tissue slides. Hematoxylin and eosin (H\&E) is a standard stain routinely employed worldwide. It is easy to acquire and cost effective, but cells and tissue components show low-contrast with varying tones of dark blue and pink, which makes difficult visual assessments, digital image analysis, and quantifications. These limitations can be overcome by IHC staining of target proteins of the tissue slide. IHC provides a selective, high-contrast imaging of cells and tissue components, but their use is largely limited by a significantly more complex laboratory processing and high cost.
We proposed a conditional CycleGAN (cCGAN) network to transform the H\&E stained images into IHC stained images, facilitating virtual IHC staining on the same slide. This data-driven method requires only a limited amount of labelled data but will generate pixel level segmentation results. The proposed cCGAN model improves the original network \cite{zhu_unpaired_2017} by adding category conditions and introducing two structural loss functions, which realize a multi-subdomain translation and improve the translation accuracy as well. 
Experiments demonstrate that the proposed model outperforms the original method in unpaired image translation with multi-subdomains. We also explore the potential of unpaired images to image translation method applied on other histology images related tasks with different staining techniques.

\end{abstract}

\section{Introduction}
Virtual staining, by its literal meaning, is to use computerized algorithms to create an artificial effect of staining without physically tampering the slide. With well-designed applications, this novel approach will introduce a revolutionary impact on histopathology analysis of digitized whole slide images (WSIs). 
Based on the digital scan of a real slide stained using a traditional dyeing method, virtual staining is, in fact, a post-processing step that can generate other stained versions of the same slide using dedicated computer algorithms. The aim is to provide different staining effects that can to highlight different relevant histological and cell features on the same slide.
From this point of view, image normalization can be also regarded as a basic form of virtual staining, which mainly focuses on reducing the staining variance among slides caused by different staining protocols, scanners or scanning conditions, and ensuring a coherent appearance of components.

Virtual staining can be considered in two main types: low-level virtual staining and semantic virtual staining.
In low-level virtual staining, the output image is a weighted liner combination of different channels of the original image in its colour space. There are not semantic factors considered in producing the resulting version of the image. 
In comparison, the semantic virtual staining methods further take into account semantic information in the transform process to ensure the semantic correctness or truthfulness of the virtually stained images. 

\subsection{Challenges and Objective}
The large size of unannotated histology image data has posed critical challenges for its understanding and analysis. Motivated by the great success of deep learning models applied in different tasks in natural image analysis, more and more deep learning algorithms and systems are being designed for histopathology image analysis. Dozens of fruitful outcomes have been achieved based on deep convolutional neural networks (DCNN), especially using patch-based methods. With enough number of annotated patches, a DCNN will be trained to detect cancerous tissue patches against other benign tissue components. 

\begin{figure*}[t!]
    \centering
    \begin{subfigure}[t]{0.5\linewidth}
        \centering
        \includegraphics[width=0.5\linewidth]{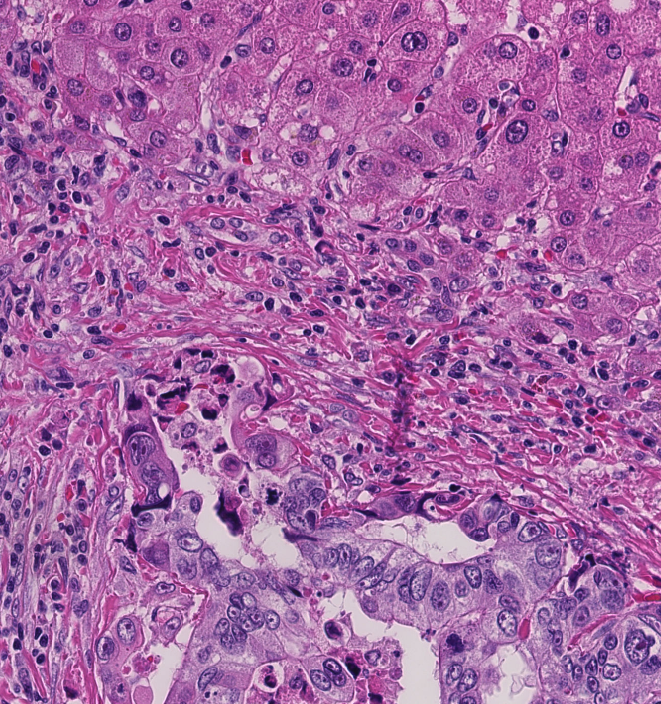}
        \caption{H\&E stained image}
    \end{subfigure}%
    \begin{subfigure}[t]{0.5\linewidth}
        \centering
        \includegraphics[width=0.5\linewidth]{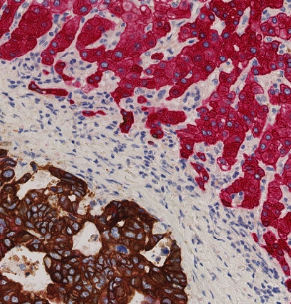}
        \caption{Immuno stained images} 
    \end{subfigure}
    \caption{Matched H\&E and IHC stained images of a similar WSI region}
    \label{fig:complex_region}
\end{figure*}

However, there are intrinsic limitations in the current methods. A major issue is networks' high dependence on annotated training sets that are often inadequate both in terms quantity, due to the expensive manpower consumption required for the job, and in terms of quality, because of inter- and intra-observer subjectivity. Moreover, when multiple tissue types are present in a region intertwined together, it becomes impossible even for experienced pathologists to create accurate annotation masks, such as the H\&E stained image region shown in Fig. \ref{fig:complex_region} (a).  The lack of adequate and accurate training data results in unsatisfactory detection accuracy.

In clinical practice, there are other methods, mainly IHC  staining, for the pathologist to highlight the cancerous or other cell and tissue components of interest. IHC staining allows the visualization of specific proteins on the tissue slide using targeted antibodies and subsequently detecting the bound antibodies using chromogens of different colours. In such a way, by using protein-binded antibodies that are selectively expressed in the cancer cells, these become readily distinguishable from the other tissues by the high-contrast chromogen staining, as shown in shown in Fig. \ref{fig:complex_region} (b).
Despite its superior features compared to H\&E, there are obstacles that preclude the wide application of IHC staining in clinical practice - the high cost of antibodies, autostainer machine equipment, and complex laboratory process. Unfortunately, these issues are unresolvable currently. In most cases, pathologists have to rely on H\&E or other relatively cheaper and more available staining methods for their diagnosis and limit the use of IHC for a very small number of selected slides. For this reason, the majority of computer models and systems for histopathological cancer grading are designed for segmenting and classifying the tumor tissue based on H\&E imaging. 

If there is a way to model the pairwise relationship between the tissue morphology in H\&E and IHC stained conditions, it is then possible to predict one from the other.
Recently, with the continuous progress of deep learning on different image processing tasks, it becomes possible to 'translate' one image to another with a different style. Such technology brings the opportunity to produce an image virtually stained with one dye from another, for example, producing IHC images from images dyed with the relatively cheaper and widely available staining techniques such as H\&E.
This may be very useful in many clinical diagnostic and AI-based applications, such as to improve the effectiveness of pathological examination by reducing the eyeballing time in visual screening of the slides; and to increase the segmentation and classification performance of AI models, i.e. if the H\&E slides can be virtually IHC stained, with a few simple post-processing, the generated IHC images can provide highly precise segmentation of tissue regions of interest.

What is more interesting, given an original image acquired with one stain, virtually stained images with one or more other dyes can now be generated from exactly the same sample. Currently, only one or two chromogens can be employed on the same sample for brightfield microscopy, which is the standard in histopathology practice. However, it is often the case that several IHC stainings need to be performed on the same tissue region to extract multiple features of diagnostic relevance. In such cases, pathologists perform serial cuts from the same tissue block, each between 3µm and 5µm thick, and stain them with different techniques and antibodies. This introduces inevitable inter-slide variability in the cell and tissue structures as, while performing serial cuts along the z-axis, neighbouring tissue regions are typically similar but do not match exactly each ocher, preventing cell-level segmentation and colocation analyses across slides. Consequently, the differently stained sections do not represent exactly the same tissue sample. With the help of virtual staining, multiplexing with two or more multiple stains on the same sample becomes possible.

There are many ways to understand or implement virtual staining. It can be considered as a special kind of style transfer that only transfers the colour coding to the original images while keeps the structural information consistent.  
In this case, it is also similar to the image colourization problem which will re-colourize the image with new colours. However, this method needs to base on the good performance of semantic segmentation.
Based on the tightly related common features of images with different stainings, the most relevant problem is the unpaired image-to-image translation that will transfer the original image with style from the reference image.

Therefore in this study, we tackle the task of virtual staining tissue samples by focusing on producing virtual IHC images from original H\&E images as a start. With the proposed novel cCGAN model, the challenging objective of unpaired image-to-image translation for multi-class virtual staining is addressed, with additional patch-wise labels.
In particular, to make sure the structural details of the original image remains unchanged in the virtual staining process, a photorealism and structure similarity loss is introduced to regulate the translation process.
The overall aims are to improve the effectiveness of pathological examination and enable a more precise tissue segmentation and labelling for further computer aided applications, eventually based on the virtual IHC images.

\subsection{Related Work}
\textbf{Generative Adversarial  Networks  }
Generative Adversarial Networks (GANs) \cite{goodfellow_generative_2014}, as an important branch of deep learning, is gaining more and more attention from the researchers around the world.  A significant amount of investigations have been made to explore the potential of GAN in natural images related tasks like image synthesizing, image super-resolution, and style transfer \cite{wang_high-resolution_2017,zhang_colorful_2016}.
The generative model is a very promising approach for histology image processing as well. The fundamental idea of a generative adversarial network is to train the generative model $G$ that can generate fake images though learning the real images, to fool the discriminator $D$. Once the discriminator $D$ can not tell the fake ones from the real, it means that the network has learned to model the distribution of the input data appropriately.
Modelling the patterns in the histology images is a particularly complicated task for generative models. Although there are underlying regulations that control the growth pattern of the cells, the regulations are limited and unknown to human experts. Tissue morphologies can be treated as orderless texture patterns. In this case, the generative adversarial networks are often used as a data augmentation method that helps generate more tissue regions. This is one of the most intuitive uses of generative models in histology image processing.
\\
\textbf{Conditional GAN / Paired Image to Image Translation}
Based on the original GAN networks, conditional GAN\cite{mirza_conditional_2014} has been proposed to take into account certain constraints that can help to improve the truthfulness of images, in additional to the imaginary information. 
The conditions encode the labels of the generated images which contain category information.
Furthermore, pix2pix GAN, as a paired image-to-image translation method, provides segmentation masks as conditions to generate images on a pixel level \cite{isola_image--image_2017}.
From the image translation point of view, the pix2pix model translates the images from abstract representation to real images while keeping the semantic meaning.\\
\textbf{Unpaired Image Translation}
Among the above mentioned paired image-to-image translational models, an obvious drawback is the requirement for corresponding masks on the input images.
Thus, a few unpaired image translation models are proposed including SimGAN \cite{shrivastava_learning_2017}, CoGAN \cite{liu_coupled_2016} and CycleGAN\cite{zhu_unpaired_2017}.  
Though the images are unpaired and are from different domains, they shared similar semantic and structural features, which form the basis for unpaired image translation.
For example, in CycleGAN, the model tried to translate a horse to a zebra and an apple to an orange. The objects before and after the translation have to demonstrate similar semantic structures to make the translation meaningful. \\
 \textbf{Style Transfer}
Instead of re-arranging the contents in the images, the style transfer approach attempts to replace the low-level representations from another domain regardless of whether the input and reference images share the same semantic information or not\cite{gatys_image_2016}.
However, the transferred images may change significantly compared to the original, in terms of both content semantics and structure.
Hence, deep photo style transfer tries to keep the structural information as much as possible while changing the semantic meaning by manipulating the colour space \cite{luan_deep_2017}.

Among the application domains for image translation, histology image analysis poses unique challenges. How to improve the visualization of histopathology images is at the core of the challenge due to its crucial impact on the efficiency of pathology examination and diagnostic accuracy.
Research on virtual staining has been conducted for many years and applied on many different types of images. The recent advances in GAN based approaches with their superior abilities open new roads in this direction.\\
\textbf{Low Level Virtual Staining}
Early virtual staining research mainly focuses on the low visual level.
For example, in the work of  Sasajima et al.,  a real-time endoscopy system is developed to visualize the endocytoscopic images in a H\&E staining style \cite{sasajima_real-time_2006}.
In Bautista and Yagi's work, they present a linear spectral transformation method for  "digital staining of histopathology multispectral images" \cite{bautista_digital_2012} .
Other works like in \cite{elfer_draq5_2016,giacomelli_virtual_2016}, employ a simpler mapping method to virtually stain the fluorescence images to H\&E.
Some other research attempt to modify the hardware to achieve the virtual staining results.
Tao et al. present a non-linear microscope to assist the diagnosis of breast cancer \cite{tao_assessment_2014}.
Recent work has demonstrated that with the help of deep learning, the different tissue components can be separated semantically. This means that the staining of different tissue components can be separately re-colourized regardless of the original staining technique.
Bayramoglu et al. utilise a Conditional Generative Adversarial Networks (cGAN) to virtually stain unstained specimens \cite{bayramoglu_towards_2017}.
Rivenson et. al\cite{rivenson_deep_2018} also employ the GAN model to virtually stain the fluorescence images to H\&E images.\\
\textbf{Semantic Virtual Staining}
Generally speaking, semantic virtual staining can be considered as a deeper process of raw images which is usually related to image segmentation or classification tasks. For multi-class tissue type classification and segmentation, once the results are mapped back to the original images, the new results can be a form of virtual staining. The research by Litjens et al. overlapped a heatmap of the likelihood of cancerous regions on the original image, which is also a kind of virtual staining \cite{litjens_deep_2016} .
Recently, a team from Google successfully build an augment reality microscope which has built-in "virtual staining function" through real-time image analysis using deep learning  \cite{pohsuan_augmented_nodate}.
Trahearn et al. \cite{trahearn_hyper-stain_2017} propose a hyper stain inspector for image alignment and cancer detection as well.
Another innovative way of "virtual staining" method is proposed in \cite{tosun_histological_2017}. The authors represent different tissue components in the image with colourful plates of different size, facilitating diagnosis.

\subsection{Our contribution}
In this study,  we explore the potential of unpaired image-to-image translation as "virtual staining" for histopathology image analysis. We propose a conditional CycleGAN which could perform multi-class virtual staining with additional patch-wise labelling. Furthermore, to make sure the virtual staining does not change the structural details of the original image, we introduce photorealism and structure similarity losses to regulate the translation process.

\section{Method}

\begin{figure}[h]
\centering
\includegraphics[width=0.4\linewidth]{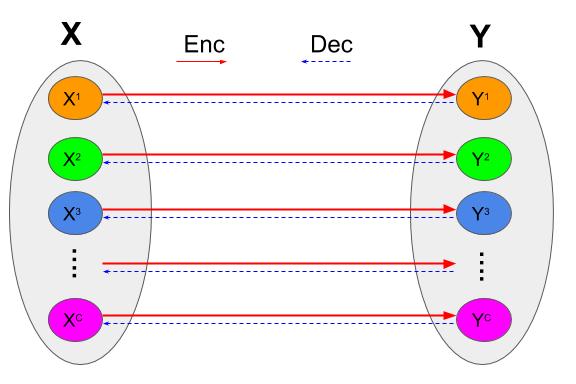}
\caption{Mapping between two domains and their sub-domains}
\label{fig:mapping}
\end{figure}

The overall objective of the proposed network is to learn a multi-class mapping between two domains $X$ and $Y$. In fact, the transformation takes place between their sub-domains  $X^{t}$ and ${Y^{t}}$ ,$t\in [1,C]$, where $C$ denotes the number of predefined classes.  For two unpaired samples $x $ and $y$   , their distributions of the training dataset  are denoted as $x \sim p({x|c})$ and $y \sim p(y|c)$ where $c$ represents the category condition.
As illustrated in Fig. \ref{fig:mapping},  there are two conditional mappings in our model, the encoder network $Enc$  maps the input image from domain  $X$ to  $Y$  while the decoder network $Dec$ maps back from $Y$ to $X$.

Our method extends the original CycleGAN model proposed in \cite{zhu_unpaired_2017} for the histology image tasks, by introducing the following two improvements:
\begin{itemize}
\item Patch-wise label information is employed to enforce the model to learn mutual representation with conditional query. This is achieved by appending a few classifier layers to the original generative network for the sub-domain consistency.
\item Two constraints are introduced to regularize changes to the structural details: the photorealism loss and the structural similarity loss.
\end{itemize}
\begin{figure}[h]
\includegraphics[width=\linewidth]{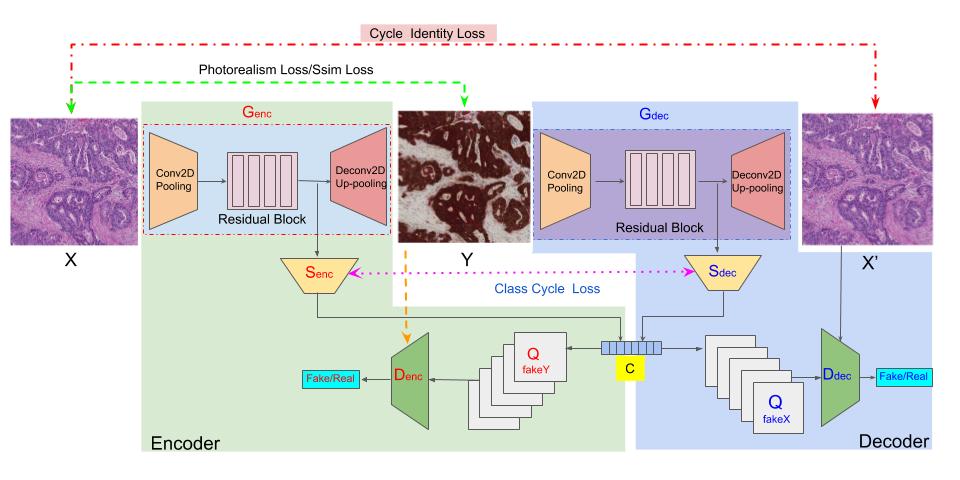}
\caption{Overall structure of the proposed network}
\label{fig:framework}
\end{figure}
As shown in Fig. \ref{fig:framework},  the whole framework consists of two big modules- an encoder network (in green zone) and a decoder network(in blue zone), both of which have the same subnetwork structure, a generator $G$,  a discriminator $D$ and a classifier $S$. 
In the encoder network, they are denoted as $G_{enc}$, $D_{enc}$ and $S_{enc}$. Likewise, in the decoder network, they are $G_{dec},D_{dec}$ and $S_{dec}$.
The generator is in charge of translating the image between domains, while the discriminator provides feedback on True or False for the generated image. At the same time, the classifiers will provide extra information about the image's sub-domain.
To train the proposed model for decent performance on the task, we facilitate the following losses during the training phase.

\subsection{Conditional Adversarial Loss}

To include the category information for guiding the translation process,  we apply conditions on the inputs for the discriminator $D_{enc}, D_{dec}$.
For the mapping from $X$ to $Y$, the conditional adversarial loss inspired by the loss in\cite{goodfellow_generative_2014} within the class $c$ and the generator $G_{enc}$ is defined as :
\begin{equation}
\begin{aligned}
{\mathcal{L}}_{cGAN}(G_{enc},D_{enc})  = \mathbb{E}_{y\sim p(y|c)}[logD_{enc}(y)]\\
+\mathbb{E}_{x\sim p(x|c)}[log(1-D_{enc}(G_{enc}(x)))]
\label{eq: cond_adloss}
\end{aligned}
\end{equation}
Likely, for the mapping from $Y$ to $X$,  the same loss is calculated with the discriminator $D_{dec}$ and generator $G_{dec}$.

\subsection{Deterministic Loss for Tissue Classification Networks}
The tissue subtype is a kind of crucial information for the histopathology image translation.  Without appropriate class inference, it will lead to a failed translation.
In our network, the classifiers $S$  infer the category information independently of the generator but share the same base network structure.
Practically, by adding extra category information to the network, we attempt to map the distribution of the data category information during the transformation, and include the categories information in the output as well.
The networks employ the Softmax Entropy Loss $\ell$ to regularize the category information  with regards to their class information. The loss of the classifiers is defined as :
\begin{equation}
{\mathcal{L}}_{class}(S_{enc},S_{dec})  = \mathbb{E}_{x\sim p(x|c)}[\ell(S_{enc}(x),c)] + \mathbb{E}_{y\sim p(y|c)}[\ell(S_{dec}(x),c)] 
\label{eq: loss_class}
\end{equation}
\subsection{Cycle Loss and Classification Cycle loss}
The core idea of cycle-GAN is to use the encoder and decoder process as a cycle to make $X \approx X'$ after one cycle $X \rightarrow Y \rightarrow X^{'}$.  By minimizing the difference between $X$ and $X^{'}$, the network will be able to learn the shared features.
The original cycle loss defined as \cite{zhu_unpaired_2017} :
\begin{equation}
\begin{aligned}
\label{eq:loss_cycle}
\mathcal{L}_{cyc}(G_{enc},G_{dec})  = \mathbb{E}_{x\sim p(x|c)}[||G_{dec}(G_{enc}(x))-x||_{1}] \\
                                                                            +\mathbb{E}_{y\sim p(y|c)}[||G_{enc}(G_{dec}(y))-y||_{1}]                                   
\end{aligned}
\end{equation}
During the cycle process, the outputs of the classifiers $S$ should be identical as well.  Thus the classification cycle loss is defined with L1 loss:
\begin{equation}
\mathcal{L}_{clcyc}(S_{enc},S_{dec})  = \mathbb{E}_{(x,y)\sim p(x,y|c)}[||S_{enc}(x)-S_{dec}(y)||_{1}] 
\end{equation}
The cycle consistency of the classification information is of signification importance to final results.
\subsection{Identity Loss , Photorealism Loss and  Structural Similarity Loss}
The identity loss proposed by Taigmen et al. \cite{taigman_unsupervised_2016}
is introduced  in the original cycle GAN paper to preserve  the color composition  especially when the input image is very close to the output image domain\cite{zhu_unpaired_2017} . The definition of identity loss for the proposed network is demonstrated as follow:
\begin{equation}
\begin{aligned}
\mathcal{L}_{id}(G_{enc},G_{dec})  = \mathbb{E}_{x\sim p(x|c)}[||(G_{enc}(y)-x||_{1}] \\
                                                                            +\mathbb{E}_{y\sim p(y|c)}[||G_{dec}(y)-y||_{1}] 
\end{aligned}
\end{equation}
However, in virtual staining, identity loss is not strong enough to keep  the structural information unchanged, because the mapping is on the same domain.
Hence, we introduced two other loss functions, photorealism loss \cite{luan_deep_2017}  and structural similarity loss (SSIM)\cite{wang_image_2004}, to enhance the original cycle-GAN network. 
These two losses serve the same purpose which is to preserve the texture structure of the input image.
Photorealism loss uses  Matting  Laplacian transform to measure the structural differences and is defined as :

\begin{equation}
Pho(x,y) = \sum_{k=1}^{3}y_{k}^{T}[M_{x}]y_{k} +\sum_{k=1}^{3}x_{k}^{T}[M_{y}]x_{k} 
\end{equation}
where $k$ denotes the RGB channels while $M_{x}$/$M_{y}$ is the  transformed matrix regarding to the input images  $x/y$.
For more details, please refer to \cite{luan_deep_2017}.

\begin{equation}
\mathcal{L}_{pho}(G_{enc},G_{dec}) =  Pho(G_{enc}(x),x) +Pho(G_{dec}(y),y)
\end{equation}

SSIM has been used for assessing the image quality in many related studies \cite{wang_multiscale_2003} .
We introduce it to our model to regulate the structural changes between the input and output images.
For each pixel  in the image, the SSIM is defined as:
\begin{equation}
\begin{aligned}
Ssim(x,y) & = \dfrac{2\mu_{x}\mu_{y}+Q_{1}}{\mu_{x}^{2}+\mu_{y}^2+Q_{1}} + \dfrac{2\sigma_{xy}+Q_{2}}{\sigma_{x}^{2}+\sigma_{y}^{2}+Q_{2}} 
\label{eq:ssim}
\end{aligned}
\end{equation}
where $\mu_{x},\mu_{y}$ are the mean of a fixed window centered as the pixel, $\sigma_{x},\sigma_{y}$ are the standard derivations.  $Q_{1},Q_{2}$ are the regularization term for division stabilization. 
Hence, the loss function of SSIM of the whole network can be formulated as:
\begin{equation}
\mathcal{L}_{sim}(G_{enc},G_{dec}) = (1-Ssim(G_{enc}(x),x)) +(1-Ssim(G_{dec}(y),y))
\end{equation}

\subsection{Our Approach}

Our full objective can be  achieved by a weighted linear combination of: \\
\begin{equation}
\label{eq:final_loss}
\begin{aligned}
\mathcal{L}(G,D,S,x,y,c) = {\mathcal{L}}_{cGAN}(G_{enc},D_{enc})   
                                                                +{\mathcal{L}}_{cGAN}(G_{dec},D_{dec})   \\
                                                                +\lambda{\mathcal{L}}_{cyc}(G_{enc},G_{dec}) 
                                                                 +\delta{\mathcal{L}}_{id}(G_{enc},G_{dec})\\
                                                                +\gamma{\mathcal{L}}_{class}(S_{enc},S_{dec})
                                                                +\gamma{\mathcal{L}}_{clcyc}(S_{enc},S_{dec})\\ 
                                                                +\alpha{\mathcal{L}}_{ssim}(G_{enc},G_{dec}) 
                                                                +\beta{\mathcal{L}}_{pho}(G_{enc},G_{dec}) 
\end{aligned}
\end{equation}
The parameters $\lambda,\gamma,\delta,\alpha,\beta$ in the loss function regulate the importance of different losses to the overall objective.
By solving the following equation, 
\begin{equation}
G,D,S = \arg {\min\limits_{G,S}{\max\limits_{D}}{\mathcal{L}}(G,D,S,x,y,c)}
\end{equation}
optimal models can be found for the generators $G_{enc},G_{dec}$ , the decoders $D_{enc},D_{dec}$ and the classifiers $S_{enc},S_{dec}$.

\begin{figure}[!h]
\includegraphics[width=\linewidth]{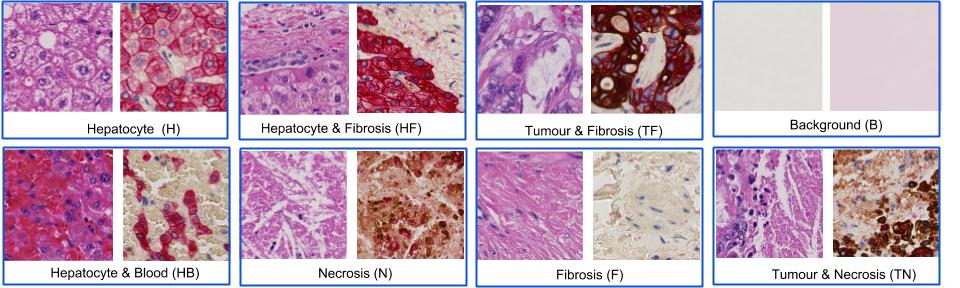}
\caption{Example of the training datasets}
\label{fig:trainingset}
\end{figure}

\section{Dataset and Implementation}
\subsection{Dataset preparation}
Our dataset consists of whole slide images from resected colorectal liver metastases (CRLM) operated at **(hospital name)**. The study was approved by the **(ethical approval number)**. Matched H\&E and IHC stained slides (2-plex CK19/CK18) were scanned at 40X  with Hamamatsu NanoZoomer slide scanner. On H\&E stain, all cell types show varying tones of dark blue and pink. CK19 stains in brown colorectal cancer cells using DAB chromogen, while CK18 stains in red the surrounding and benign liver cells (hepatocytes) with AP chromogen. Immune cells, stromal cells and extracellular matrix (fibrosis) are negative for both cytokeratine (CK) stainings. Immunohistochemical sections are also counterstained with hematoxylin (\textit{H}) to visualize the background context of cells and tissue components on the slide.

As histological sections are 4 $\mu m$ thick, there is always some degree of inter-slide variation between the matched slides. The differences may not be apparent at low magnification levels, for example $10\times$, but become evident at higher magnification levels, e.g. $40 \times$. In this study, the dataset are cropped from $20 \times$ magnification level with a patch size of $256 \times 256$ pixels.

According to the color properties of the IHC stained image, for the training dataset we divide the training patches into 8 different categories. As demonstrated in Fig.\ref{fig:trainingset}, they are \textit{Hepatocyte (H), Fibrosis (F), Necrosis (N), Tumour \& Fibrosis (TF), Hepatocyte} \& \textit{Fibrosis (HF), Hepatocyte} \& \textit{Blood (HB), Tumour} \&\textit{ Necrosis (TN) and Background (BG)}. The testing dataset  are cropped from  aligned WSIs with a patch size of $256 \times 256$ pixels. 
The number of patches that are used for training are listed in Table \ref{tab:quantitive}.

\subsection{Implementation Details}
\textbf{Network Architecture}
This section describes the implementation details including the network structure and the parameters set-up.
The generator networks have the same architecture as proposed in \cite{johnson_perceptual_2016}.  Two convolutional layers at the begining for downsampling and two deconvolutional layers for upsampling. In the middle, there are 9 resnet blocks \cite{he_deep_2016}. Both the discriminators and classifiers are composed in  a fully convolutional fashion. For the discriminators, there are  5 layers  inside, while for the classifiers, the number of convolutional layers are 8.\\

\noindent
\textbf{Training Details}
During the training, the history of generated images is used to reduce model oscillation. However, to fit with the conditional generative network, the image query process is applied on condition as well. For the parameters $\lambda,\gamma,\delta,\alpha,\beta$  in the loss functions Eq.\ref{eq:final_loss}, $\lambda$ is set to a fixed value of 10, $\gamma$ is set to 0.5 and the other three ($\delta, \alpha, \beta$) are set to different values to compare the performances. The rest of the parameters are identical with the original CycleGAN.
 \begin{figure}[h]
\centering
\includegraphics[width=\linewidth]{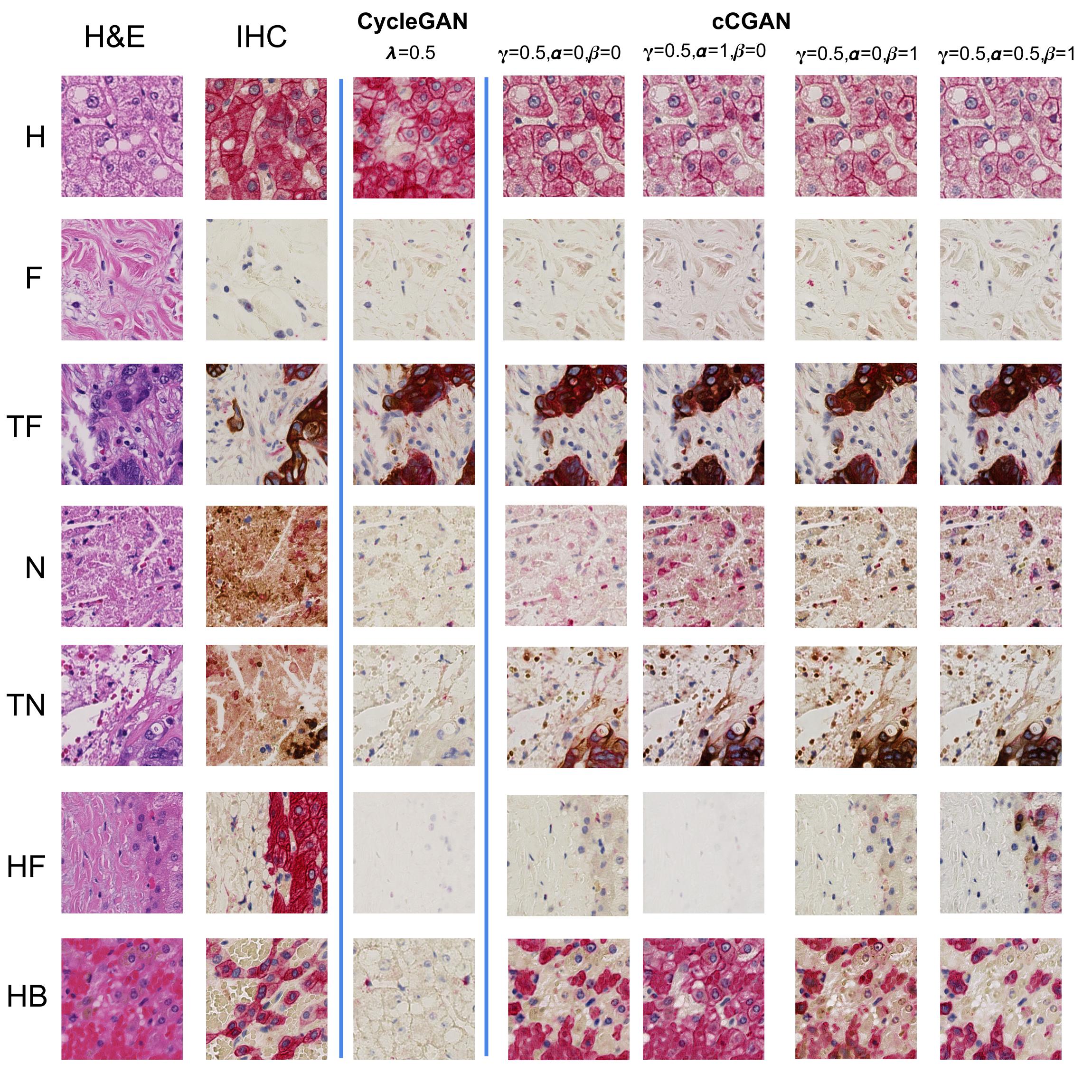}
\caption{Experiment results with different parameters settings}
\label{fig:result_v1}
\end{figure}
\section{Results and Evaluation}
Evaluating the results of the generated images for tasks like style transfer, image re-colorization is well-known  challenge for  a long time.
One of the wide adopted methods is to use Amazon Mechanical Turk (AMT)  perceptual study that asks human observers to differentiate the fake from the true ones. However, in our study, we need to employ well-trained histopathologist experts to distinguish them. Meanwhile, we will ask computer vision researchers to assess on the image by a visual quality error score. 

For the computer vision researchers, all the patches (real and fake) from different classes are mixed together and the observers need to make decisions based on the image quality. While, for professional assessment, both of the original H\&E stained  and virtually stained patches will be provided. The experts will give a  staining score that indicates if the patches have been properly virtual stained. Two pathology experts and ten computer vision researchers are invited to perform the evaluation. 240 patches are generated  from the evaluation dataset (30 patches per class). The results of experts are listed in Table \ref{tab:quantitive}.
The results of different settings are demonstrated in  Fig. \ref{fig:result_v1}. In this figure, for each H\&E stained patch, the patch from the same location on the  roughly aligned IHC images are used as references.
Comparing to the results by CycleGAN, the results of proposed method greatly improve the results especially in the class with mixed components and limited training examples such as \textit{TN, HF, HB}.
From the experiments that we can observe the $\gamma =0.5,\alpha =0.5$ and $\beta =1$ output the best results. Hence, we employ the results from this model for  further evaluations.
Further results are listed in Table \ref{tab:quantitive}.
\begin{table}[h]
\centering
\caption{Quantitative Evaluation Results}
\label{tab:quantitive}
\begin{tabular}{p{4.3cm}|p{0.7cm}p{0.7cm}p{0.7cm}p{0.7cm}p{0.7cm}p{0.7cm}p{0.7cm}p{0.7cm}|p{0.7cm}}
\hline
                        & H    & TF  & N   & F    & HF  & TN & HB  & BG   & Overall \\ \hline
No. Annotated patches (H\&E)    & 1828 & 950 & 538 & 798  & 212 & 48 & 394 & 2016 &3392 \\
No. Annotated patches(IHC)     & 1440 & 738 & 438 & 1184 & 210 & 52 & 88  & 1252 &2711\\
No. Test   patches  (H\& E)           & 30   & 30  & 30  & 30   & 30  & 30 & 30  & 30   & 240\\
Vision experts on CycleGAN* &   3\%   & 26\%    &  1\%   & 4\%     & 3\%    & 0\%  & 18\%    &0\% & 7\%     \\
Vision experts on cCGAN*   &  0\%    & 10\%    &  4\%    &  0\%    & 0\%    & 0\%   & 0\%    &0\% &  2\%   \\
Pathologists on CycleGAN*  &   34\%   &  71\%   &  22\%   & 11\%     & 83\%    &  95\%  & 97\%    & 0\%     & 52\%   \\
Pathologists on cCGAN*  & 23\%    & 39\%   &   23\%  & 4\%     & 76\%    & 86\%   & 99\%    & 0\%   &44\%   \\ \hline
\end{tabular}\\
* The number indicates the visual quality error score and staining error score. The larger the number indicate a worse performance.
\end{table}

\begin{figure}[h]
\centering
\includegraphics[width = 0.75\linewidth]{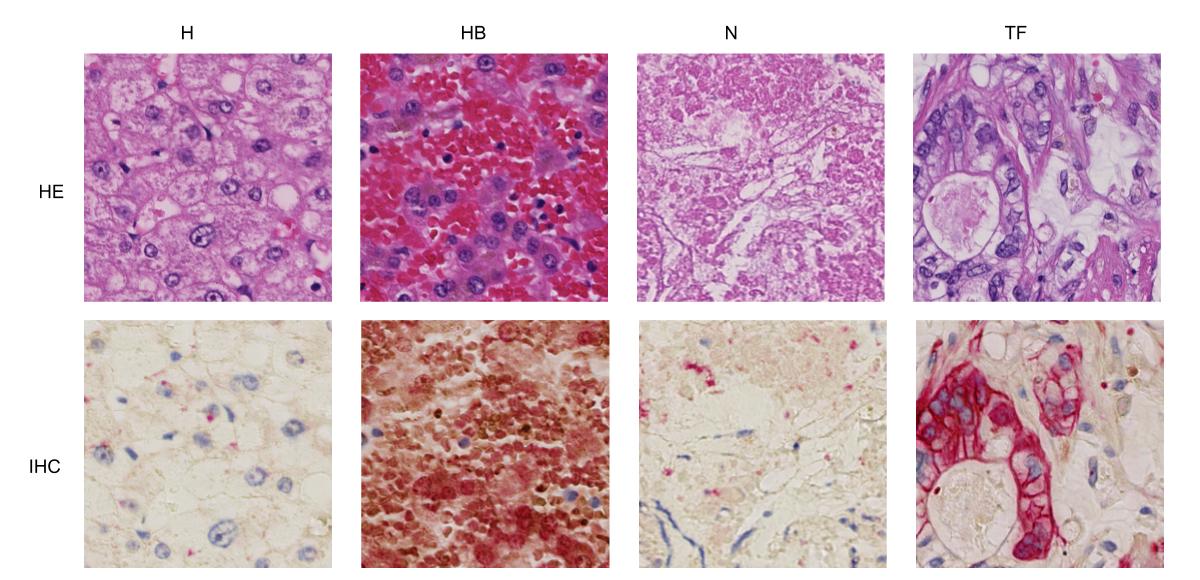}
\caption{Unsuccessful mapping examples}
\label{fig:fail_eg}
\end{figure}

The results from the non-experts view demonstrated that the proposed method is better to preserve the image contents than the original cycleGAN. The introduction of the structural losses to the model greatly suppresses the "imaginary" ability, especially when encountering new features. The class \textit{TF} has the highest fake rate, which has a strong relationship with its complex features.
 The poor performance in the classes \textit{HB, HF, and TN} is due to the insufficient and unbalanced training data in different classes. 
By inspecting the failed transformation cases, we can find that mapping to the wrong domain is the main cause for the failure. 
Some examples are illustrated in Fig. \ref{fig:fail_eg}. The patch from H is mapped to F, while for the HB, N, and TN, the staining style is N, F and H respectively.

\section{Conclusions and Future Work}
The results indicate that the proposed virtual staining method is capable of virtually staining the H\&E images into IHC style image based on the underlying mutual features in the two kinds of images. In other words, IHC stained WSI are distinguishable enough for guiding the cross-domain translation.

With the help of virtual staining, the laborious and expensive experts' manual annotation can be minimized. In fact, the presented method for computerized virtual staining entails great potential for a wide range of clinical diagnostic and AI related applications. A few of these are listed below:
\begin{enumerate}
\item Fast and low-cost generation of IHC staining for improved tumour diagnosis. Currently, pathologists devote a great amount of time and efforts in the error-prone eyeball work of searching for tumour cells within the low-contrast H\&E slides. Virtual IHC of tumour cells, by enabling high-contrast visualization, may contribute to greatly reducing visual screening time and improving the accuracy of cancer detection. 
\item Virtual multiplex IHC staining for spatial mapping and objective quantitation of intertwined cell types. As a generalization of GAN networks can be trained to classify and virtually stain multiple cell types of interest, provided multiplex immunostainings as training. For example, the spatial analysis and quantitation of tumour and immune cells is a crucial assessment in the current cancer immunotherapy. IHC stained cells can be then readily quantified using available digital pathology image analysis tools.
\item WSI registration for colocation analysis in multiple IHC stainings. Registration of multiple serial sections is another challenge of increasing interest in the diagnosis and biomedical research. In clinical practice, the pathologists employ multiple IHC stainings (bio-markers) to detect relevant features in cells and tissues within the WSIs.
As it is often not feasible to align the samples during lab processing, image registration is the only option to accomplish this task, but very difficult to achieve.
By exploiting virtual staining's high accurate matching of image pairs, multiple aligned IHC stainings can be combined into a new staining style achieving increasing levels of multiplex immunostaining far beyond current laboratory techniques.
\item Automatic generation or augmentation of highly accurate training datasets. As, once learned, virtual staining can be exploited bi-directionally, H\&E staining can also be faithfully reproduced from an IHC stained slide as demonstrated in Fig. \ref{fig:hic2he}. The original, high-contrast IHC staining can then be easily segmented and used as mask to automatically extract annotations from the virtual H\&E staining with a pixel-level accuracy, reducing the need for manual annotation and overcoming the intrinsic limitation of inter-slide variation in serial (4 $\mu m$ thick) tissue slides. This may enable also to improve the accuracy of patch-based training data, especially in areas with a mixture of cell  and tissue types.



\item Color deconvolution and nuclei detection. Virtual staining can also be used to decouple the information of the H\&E stained image to obtain only the H channel of the image. By using H\&E, H-only, and E-only stained WSIs as training data, the model will learn to "unmix" the two color components in the H\&E images. The H/E-only images can then be easily normalized. This may facilitate nuclei detection tasks, especially in cell dense and cancerous areas. Some preliminary results are demonstrated in Fig. \ref{fig:deconv}. \\
\end{enumerate}

\noindent
\begin{minipage}{0.48\linewidth}
\centering
\includegraphics[width=\linewidth]{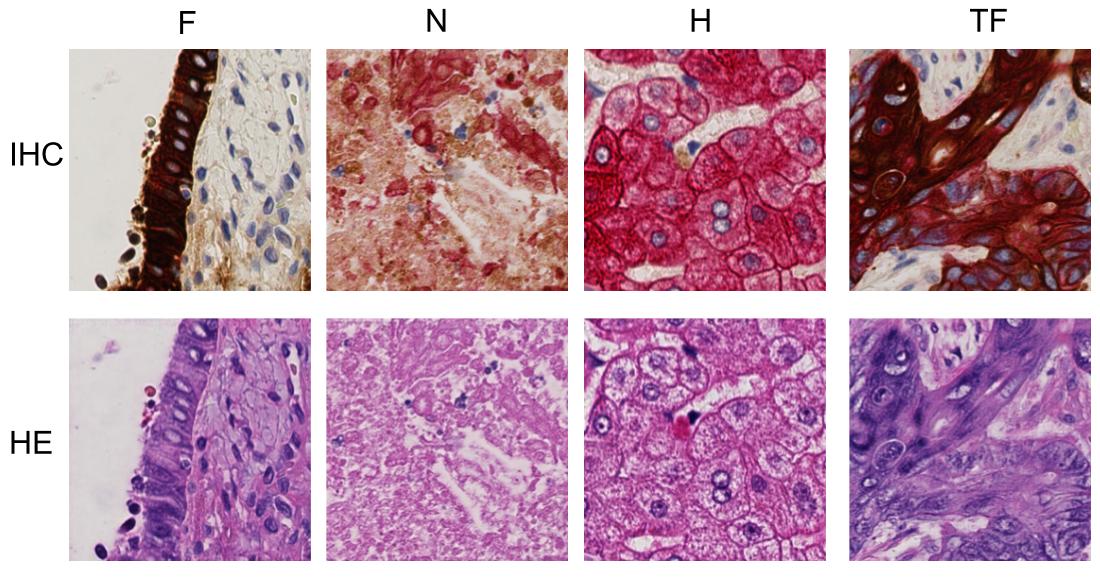}
\captionof{figure}{IHC  to H\&E  examples}
\label{fig:hic2he}
\end{minipage}
\begin{minipage}{0.04\linewidth}
\end{minipage}
\begin{minipage}{0.48\linewidth}
\centering
 \includegraphics[width=\linewidth]{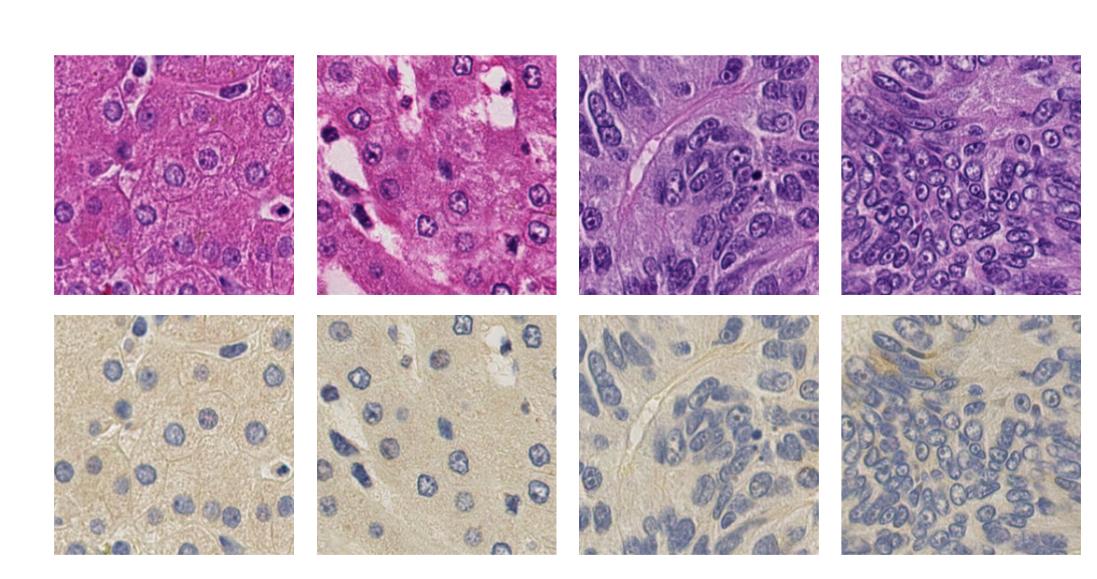}
\captionof{figure}{Color deconvolution examples}
\label{fig:deconv}
\end{minipage}\\

One of the limitations of the current study is the low number of samples for training and the significant imbalance on the amount of annotated patches among different classes. In future work, we attempt to minimize or discarding the labelling information that is used as input in order to pursue unsupervised image translation. To that end, we plan to model the staining variance within staining methods by introducing other controllable parameters.

\bibliographystyle{splncs}
\bibliography{sample}

\begin{thebibliography}{10}

\bibitem{zhu_unpaired_2017}
Zhu, J.Y., Park, T., Isola, P., Efros, A.A.:
\newblock Unpaired image-to-image translation using cycle-consistent
  adversarial networks, ({IEEE})  2242--2251

\bibitem{goodfellow_generative_2014}
Goodfellow, I., Pouget-Abadie, J., Mirza, M., Xu, B., Warde-Farley, D., Ozair,
  S., Courville, A., Bengio, Y.:
\newblock Generative adversarial nets.
\newblock In Ghahramani, Z., Welling, M., Cortes, C., Lawrence, N.D.,
  Weinberger, K.Q., eds.: Advances in Neural Information Processing Systems 27.
\newblock (Curran Associates, Inc.)  2672--2680

\bibitem{wang_high-resolution_2017}
Wang, T.C., Liu, M.Y., Zhu, J.Y., Tao, A., Kautz, J., Catanzaro, B.:
\newblock
\newblock (High-resolution image synthesis and semantic manipulation with
  conditional {GANs})

\bibitem{zhang_colorful_2016}
Zhang, R., Isola, P., Efros, A.A.:
\newblock Colorful image colorization.
\newblock In: Computer Vision – {ECCV} 2016. Lecture Notes in Computer
  Science.
\newblock (Springer, Cham)  649--666

\bibitem{mirza_conditional_2014}
Mirza, M., Osindero, S.:
\newblock
\newblock (Conditional generative adversarial nets)

\bibitem{isola_image--image_2017}
Isola, P., Zhu, J.Y., Zhou, T., Efros, A.A.:
\newblock Image-to-image translation with conditional adversarial networks,
  ({IEEE})  5967--5976

\bibitem{shrivastava_learning_2017}
Shrivastava, A., Pfister, T., Tuzel, O., Susskind, J., Wang, W., Webb, R.:
\newblock Learning from simulated and unsupervised images through adversarial
  training.
\newblock In: The {IEEE} {Conference} on {Computer} {Vision} and {Pattern}
  {Recognition} ({CVPR}). Volume~3. (2017) ~6

\bibitem{liu_coupled_2016}
Liu, M.Y., Tuzel, O.:
\newblock Coupled generative adversarial networks.
\newblock In: Advances in neural information processing systems. (2016)
  469--477

\bibitem{gatys_image_2016}
Gatys, L.A., Ecker, A.S., Bethge, M.:
\newblock Image style transfer using convolutional neural networks.
\newblock In: Computer {Vision} and {Pattern} {Recognition} ({CVPR}), 2016
  {IEEE} {Conference} on, IEEE (2016)  2414--2423

\bibitem{luan_deep_2017}
Luan, F., Paris, S., Shechtman, E., Bala, K.:
\newblock Deep photo style transfer, ({IEEE})  6997--7005

\bibitem{sasajima_real-time_2006}
Sasajima, K., Kudo, S.e., Inoue, H., Takeuchi, T., Kashida, H., Hidaka, E.,
  Kawachi, H., Sakashita, M., Tanaka, J., Shiokawa, A.:
\newblock Real-time in vivo virtual histology of colorectal lesions when using
  the endocytoscopy system.
\newblock (\textbf{63})  1010--1017

\bibitem{bautista_digital_2012}
Bautista, P.A., Yagi, Y.:
\newblock Digital simulation of staining in histopathology multispectral
  images: enhancement and linear transformation of spectral transmittance.
\newblock (\textbf{17})  056013

\bibitem{elfer_draq5_2016}
Elfer, K.N., Sholl, A.B., Wang, M., Tulman, D.B., Mandava, S.H., Lee, B.R.,
  Brown, J.Q.:
\newblock {DRAQ}5 and eosin (‘d\&e’) as an analog to hematoxylin and eosin
  for rapid fluorescence histology of fresh tissues.
\newblock (\textbf{11})  e0165530

\bibitem{giacomelli_virtual_2016}
Giacomelli, M.G., Husvogt, L., Vardeh, H., Faulkner-Jones, B.E., Hornegger, J.,
  Connolly, J.L., Fujimoto, J.G.:
\newblock Virtual hematoxylin and eosin transillumination microscopy using
  epi-fluorescence imaging.
\newblock (\textbf{11})  e0159337

\bibitem{tao_assessment_2014}
Tao, Y.K., Shen, D., Sheikine, Y., Ahsen, O.O., Wang, H.H., Schmolze, D.B.,
  Johnson, N.B., Brooker, J.S., Cable, A.E., Connolly, J.L., Fujimoto, J.G.:
\newblock Assessment of breast pathologies using nonlinear microscopy.
\newblock (\textbf{111})  15304--15309

\bibitem{bayramoglu_towards_2017}
Bayramoglu, N., Kaakinen, M., Eklund, L., Heikkila, J.:
\newblock Towards virtual h\&e staining of hyperspectral lung histology images
  using conditional generative adversarial networks, ({IEEE})  64--71

\bibitem{rivenson_deep_2018}
Rivenson, Y., Wang, H., Wei, Z., Zhang, Y., Gunaydin, H., Ozcan, A.:
\newblock
\newblock (Deep learning-based virtual histology staining using
  auto-fluorescence of label-free tissue)

\bibitem{litjens_deep_2016}
Litjens, G., Sánchez, C.I., Timofeeva, N., Hermsen, M., Nagtegaal, I., Kovacs,
  I., Kaa, C.H.v.d., Bult, P., Ginneken, B.v., Laak, J.v.d.:
\newblock Deep learning as a tool for increased accuracy and efficiency of
  histopathological diagnosis.
\newblock (\textbf{6})  srep26286

\bibitem{pohsuan_augmented_nodate}
Po­Hsuan, C., Krishna, G., Robert, M., Yun, L., Kunal, N., Timo, K., Greg~S.,
  C., Jason~D., H., Martin~C., S.:
\newblock (An augmented reality microscope for cancer detection)

\bibitem{trahearn_hyper-stain_2017}
Trahearn, N., Epstein, D., Cree, I., Snead, D., Rajpoot, N.:
\newblock Hyper-stain inspector: A framework for robust registration and
  localised co-expression analysis of multiple whole-slide images of serial
  histology sections.
\newblock (\textbf{7})  5641

\bibitem{tosun_histological_2017}
Tosun, A.B., Nguyen, L., Ong, N., Navolotskaia, O., Carter, G., Fine, J.L.,
  Taylor, D.L., Chennubhotla, S.C.:
\newblock Histological detection of high-risk benign breast lesions from whole
  slide images.
\newblock In: Medical Image Computing and Computer-Assisted Intervention
  {MICCAI} 2017. Lecture Notes in Computer Science, (Springer, Cham)  144--152

\bibitem{taigman_unsupervised_2016}
Taigman, Y., Polyak, A., Wolf, L.:
\newblock
\newblock (Unsupervised cross-domain image generation)

\bibitem{wang_image_2004}
Wang, Z., Bovik, A.C., Sheikh, H.R., Simoncelli, E.P.:
\newblock Image quality assessment: from error visibility to structural
  similarity.
\newblock IEEE transactions on image processing \textbf{13} (2004)  600--612

\bibitem{wang_multiscale_2003}
Wang, Z., Simoncelli, E.P., Bovik, A.C.:
\newblock Multiscale structural similarity for image quality assessment.
\newblock In: Signals, Systems and Computers, 2004. Conference Record of the
  Thirty-Seventh Asilomar Conference on. Volume~2., Ieee (2003)  1398--1402

\bibitem{johnson_perceptual_2016}
Johnson, J., Alahi, A., Fei-Fei, L.:
\newblock
\newblock (Perceptual losses for real-time style transfer and super-resolution)

\bibitem{he_deep_2016}
He, K., Zhang, X., Ren, S., Sun, J.:
\newblock Deep residual learning for image recognition, ({IEEE})  770--778

\end{thebibliography}

\end{document}